\documentclass{article}

 \usepackage[preprint]{neurips_2026}

\usepackage[utf8]{inputenc} % allow utf-8 input
\usepackage[T1]{fontenc}    % use 8-bit T1 fonts
\usepackage{hyperref}       % hyperlinks
\usepackage{url}            % simple URL typesetting
\usepackage{booktabs}       % professional-quality tables
\usepackage{amsfonts}       % blackboard math symbols
\usepackage{nicefrac}       % compact symbols for 1/2, etc.
\usepackage{microtype}      % microtypography
\usepackage{xcolor}         % colors

\setcitestyle{numbers}

\usepackage{algorithm}
\usepackage{makecell}
\usepackage{multicol}
\usepackage{booktabs}
\usepackage{multirow}
\usepackage{xspace}
\usepackage{textcomp}

\usepackage{verbatimbox}

\newcommand{\LLMs}[0]{\textsc{LLMs}\xspace}
\newcommand{\LLM}[0]{\textsc{LLM}\xspace}

\newcommand{\Qwen}[0]{\textsc{Qwen\-3-Co\-der}\xspace}
\newcommand{\Llama}[0]{\textsc{Lla\-ma\-3.3}\xspace}
\newcommand{\GPT}[0]{\textsc{GPT\--5.5}\xspace}
\newcommand{\Gemini}[0]{\textsc{Gemini\--3.1-Pro}\xspace}

\newtheorem{example}{Example}

\usepackage[newfloat]{minted}
\usepackage{tikz}
\usetikzlibrary{arrows,automata,backgrounds,calc,chains,external,fit,graphs,patterns,positioning,shapes,shapes.geometric}

\tikzset{
    >=stealth',
    punkt/.style={
           rectangle,
           rounded corners,
           draw=black, very thick,
           text width=10em,
           minimum height=2em,
           text centered},
    pil/.style={
           ->,
           thick,
           shorten <=2pt,
           shorten >=2pt,},
    ionode/.style={
           rectangle,
           rounded corners,
           text width=12em,
           minimum height=4em,
           text centered},
}

\tikzset{fancy/.style={rectangle,
		rounded corners=1mm,
		ultra thin,
		draw=white,
		top color=white,
		bottom color=black!20,
		draw}}

\definecolor{highClr}{rgb}{1.0, 1.0, 0.0}

\colorlet{edgeClr}{orange!80!black}
\tikzset{sandEdge/.style={
		>=stealth,
		shorten >=1pt,
		thick,
		bend left,
		text=black,
		edgeClr,
	}}

\tikzset{fadedEdge/.style={
		->,
		>=stealth,
		shorten >=1pt,
		thick,
		edgeClr!20,
	}}

\tikzset{weigthLabel/.style={
		text=black,
		sloped,
		midway,
		}}

\tikzset{fadedWeigth/.style={
		text=lightgray!40,
		sloped,
		midway,
		anchor=south,
		}}

\tikzset{blueVertex/.style={
		rectangle,minimum size=6mm,rounded corners=3mm,
		top color=white,bottom color=blue!35!cyan!25!,
		font=\ttfamily,
		text=black,
	}}

\tikzset{blueVertexG/.style={
		rectangle,minimum size=6mm,rounded corners=3mm,
		top color=white,bottom color=blue!35!cyan!25!,
		font=\ttfamily,
		text=black,
		draw=green,
		thick,
	},
	blueVertexY/.style={
		rectangle,minimum size=6mm,rounded corners=3mm,
		top color=white,bottom color=blue!35!cyan!25!,
		font=\ttfamily,
		text=black,
		draw=yellow,
		thick,
	},
	blueVertexO/.style={
		rectangle,minimum size=6mm,rounded corners=3mm,
		top color=white,bottom color=blue!35!cyan!25!,
		font=\ttfamily,
		text=black,
		draw=orange,
		thick,
	}}

\colorlet{noteClr}{lightgray!30!white!50}

\tikzset{noteBckg/.style={
		rounded corners=8pt,fill=noteClr,
	},
	noteStl/.style={
		font=\scriptsize,
		align=center,
		text=black
	}}

\title{Reliable Reasoning with Large Language Models\\via Preference-Based Maximum Satisfiability}

\author{%
  Pedro Orvalho\thanks{Work conducted while PO was a visiting researcher at the University of Oxford, under the~ELSA~Mobility~Program.}\\
  Artificial Intelligence Research Institute~(IIIA)\\
  Consejo Superior de Investigaciones Científicas~(CSIC)\\
  Barcelona, Catalonia, Spain\\
  \texttt{pedro.orvalho@iiia.csic.es}\\
  \And
  Marta Kwiatkowska\\
  Department of Computer Science\\
  University of Oxford\\
  Oxford, UK\\
  \texttt{marta.kwiatkowska@cs.ox.ac.uk}\\
  \And
  Guillem Alenyà\\
  Institut de Robòtica i Informàtica Industrial~(IRI-CSIC-UPC)\\
  Barcelona, Catalonia, Spain\\
  \texttt{galenya@iri.upc.edu}\\
  \And
  Felip Manyà\\
  Artificial Intelligence Research Institute~(IIIA)\\
  Consejo Superior de Investigaciones Científicas~(CSIC)\\
  Barcelona, Catalonia, Spain\\
  \texttt{felip@iiia.csic.es}
}

\begin{document}

\maketitle

\begin{abstract}
Large Language Models~(\LLMs) excel at understanding natural language
but struggle with optimisation tasks involving multiple constraints and user-defined preferences, which commonly arise in domains such as robotics. 
We propose a hybrid reasoning approach in which \LLMs externalise reasoning through code generation. Given a natural language problem description, an \LLM generates Python code that encodes user-defined constraints and preferences as a preference-based Maximum Satisfiability~(MaxSAT) problem, which is then solved by an exact MaxSAT solver. To ensure correctness, solutions returned by the model-generated code are independently verified for feasibility and optimality against a canonical MaxSAT encoding, allowing for different encodings and multiple optimal solutions.
We evaluate our approach using both open-source and closed-access \LLMs on three families of preference-based reasoning tasks, and compare it against direct-answer, chain-of-thought, and program-of-thought baselines using the same models.
While these baselines rarely produce feasible solutions, the MaxSAT-based pipeline achieves substantially higher acceptance rates, in some cases exceeding 80\%.
Our results demonstrate that LLM-driven code generation combined with preference-based MaxSAT enables solver-verifiable optimisation with respect to generated encodings, and substantially improves correctness under independently verified reference semantics.
\end{abstract}

\section{Introduction}

Over the past few years, Large Language Models~(\LLMs) have become integral to a wide range of everyday tasks, including writing assistance, language translation~\cite{zhu2024multilingual}, and code generation~\cite{icse24-Liang0M24}. Recent work has shown that these models can also translate natural language specifications into executable code, a paradigm often referred to as \emph{vibe coding}~\cite{sapkota2025vibe}, which can be viewed as a form of translation between natural and programming languages.
Despite these advances, the \emph{reasoning} capabilities of \LLMs remain \emph{unreliable} and difficult to verify for many reasoning-intensive tasks~\cite{shojaee2025illusion}, such as planning~\cite{guan2023leveraging,valmeekam2023planning,kambhampati2024position}, arithmetic~\cite{gao2023pal}, and logical reasoning~\cite{xu2025aristotle,calanzone2024logically,linzebralogic}. These limitations are particularly evident in optimisation problems involving multiple constraints, which frequently arise in domains such as robotics and logistics, where correctness and reliability are \emph{critical}.
Several approaches, including explicit reasoning prompts and thinking models, such as chain-of-thought~(CoT)~\cite{wei2022chain} and program-of-thought~(PoT)~\cite{DBLP:journals/tmlr/ChenM0C23} prompting, attempt to mitigate these issues. While such techniques can yield modest improvements on certain simple reasoning tasks, they provide no formal correctness guarantees. Moreover, these reasoning traces are often incorrect due to hallucinations~\cite{lin2022truthfulqa} and remain \emph{difficult to verify}, limiting their applicability to optimisation tasks that do not require rigorous constraint satisfaction and provable~optimality.

To address these shortcomings, recent work has explored hybrid neuro-symbolic approaches that combine large language models with formal reasoning tools~\cite{pan2023logic,ye2023satlm,DBLP:conf/naacl/HaoCZF25,haoplanning,shi2025constraintllm}. The core idea is to leverage \LLMs for natural language understanding and problem formulation, while delegating planning or logical inference to sound and complete symbolic solvers. Empirical results show that this division of labour can substantially improve correctness and reliability in constraint-heavy tasks.
This paradigm has been successfully applied to logical reasoning by encoding tasks directly into logical formulae~\cite{pan2023logic,ye2023satlm,shi2025constraintllm}, as well as to real-world planning scenarios, using Python and Satisfiability Modulo Theories~(SMT)~\cite{DBLP:conf/naacl/HaoCZF25,haoplanning}. Collectively, these works suggest that formalisation and verification are promising directions for overcoming the limitations of \LLMs in optimisation and planning tasks where correctness is paramount.

Furthermore, in several application domains, particularly robotics and decision-making systems, it is essential to account for \emph{user-defined preferences} in addition to hard constraints. Over the past decade, prior work has addressed preference-aware reasoning by encoding both constraints and preferences directly into Maximum Satisfiability~(MaxSAT) problems~\cite{juma2011exploiting,DBLP:conf/ai/JumaHM12,DBLP:phd/ethos/Russell12}. MaxSAT is the optimisation variant of Boolean Satisfiability~(SAT)~\cite{li2009maxsat,maxsat-handbook-sat}. While these approaches provide strong correctness and optimality guarantees, they typically require users to express their preferences in specialised languages (e.g., PDDL) or rigid formats, which require training to use.
In this context, \LLMs offer a \emph{natural interface} for capturing user intent, constraints, and preferences expressed in natural language, while logical reasoners such as MaxSAT solvers provide \emph{reliable}, and \emph{verifiable} solutions to complex multi-constraint optimisation problems. This complementary relationship motivates combining the strengths of both methods.

In this paper, we study the use of LLMs as natural language interfaces to weighted partial MaxSAT solvers for preference-aware optimisation. Our contribution is not a new generic solver-in-the-loop architecture, but an empirical and methodological study of natural-language-to-MaxSAT encoding, including canonical verification of feasibility and optimality under weighted soft constraints.
% In this paper, we propose a novel neuro-symbolic approach that integrates \LLMs with MaxSAT solvers by translating natural language optimisation problems into Python-based MaxSAT encodings, enabling reliable and verifiable \emph{preference-aware optimisation}. 
Given a natural language description of an optimisation problem, including constraints and user-defined preferences, we first prompt an \LLM to reason about and plan an encoding of the problem as a MaxSAT instance. Based on this plan and the original description, the \LLM then generates Python code using the \texttt{PySAT} API~\cite{imms18-PySAT}, which provides access to a range of SAT and MaxSAT solvers. The generated code is executed to obtain a candidate solution, which we subsequently \emph{verify} for feasibility and optimality against a \emph{canonical} MaxSAT encoding, allowing for alternative encodings and multiple optimal solutions. When a solution is infeasible or sub-optimal, structured feedback is provided to the \LLM, which is then prompted to revise its implementation.

Our empirical evaluation demonstrates that prompting \LLMs to encode optimisation problems as MaxSAT instances via Python code generation leads to substantial improvements in solution feasibility and optimality across multiple problem families, compared to relying on these models' internal reasoning alone. The results further show that problem structure, preference variety, and intermediate planning all influence performance, highlighting the importance of externalising optimisation and verification to MaxSAT solvers for robust, preference-aware reasoning.

In summary, this paper makes the following contributions.

\begin{itemize}
  \item We present a neuro-symbolic approach in which \LLMs translate natural-language constraints and preferences into MaxSAT formulations via \LLM-generated Python code, enabling formal, optimisation-based reasoning.
  \item Our method combines the flexibility of \LLMs with MaxSAT solvers to compute \emph{optimal solutions} for preference-based reasoning problems.
  \item We evaluate \LLM-generated solutions through independent feasibility and optimality verification against a \emph{canonical encoding}, without relying on the model’s internal reasoning.
  \item We analyse the impact of different user-defined preference configurations on \LLM reasoning performance across different constraint optimisation problems.
  \item Our MaxSAT-based pipeline achieves substantially higher acceptance rates than baseline approaches, in some cases exceeding~80\%.
  % \item Our code and results will be made publicly available on \textsc{GitHub} (see supplementary material).
\end{itemize}

\section{Motivation}
\label{sec:motivation}

Consider the following scheduling task. Six computational jobs must be scheduled on a single machine over discrete time slots \(\{0..7\}\). Each job must start exactly once, and at most one job may run at any time. Furthermore, the schedule must respect a set of precedence constraints: \(J_3\) must be executed before \(J_1\), \(J_5\) before \(J_0\),  \(J_5\) before \(J_4\), and \(J_2\) before \(J_1\). 
Each job is associated with a preferred deadline reflecting its relative importance. Specifically, \(J_0\) has a high-priority deadline of at most time~4; \(J_1\) has a high-priority deadline at time~0; and \(J_2\) has a high-priority deadline at time~5. The remaining jobs have medium-priority preferences: \(J_3\) has a deadline at time~2; \(J_4\) has a deadline at time~3; and \(J_5\) has a deadline at time~3. The optimisation objective is to \emph{minimise} the total penalty incurred by violating user-defined deadlines, computed as the sum of their associated weights. If multiple schedules achieve the same minimum penalty, a deterministic tie-breaking rule selects the lexicographically earliest schedule.
This problem is representative of many real-world scheduling scenarios: users naturally express constraints and preferences in intuitive terms, for instance, ``this job should finish early'', or ``that job is preferably run later'', while the underlying optimisation problem involves multiple interacting constraints and strict optimality requirements.

When prompted with chain-of-thought (CoT) reasoning, or asked to directly produce a schedule, modern \LLMs such as \Gemini, \GPT, \Llama, and \Qwen generate plausible-looking solutions accompanied by detailed reasoning traces. However, these solutions are often infeasible or sub-optimal. In this example, \GPT fails to identify a schedule that simultaneously satisfies all hard constraints and minimises the total weighted penalty, even when explicitly encouraged to reason step by step. The model returns the following infeasible solution:
\[
\{J_0:4, J_1:2, J_2:5, J_3:1, J_4:6, J_5:0\},
\]
which violates the given hard constraints, i.e., \(J_2\) before \(J_1\).
This failure does not stem from a lack of understanding of the natural language description, but from the nature of optimisation itself, as the reasoning capabilities of \LLMs are not sufficiently robust for reasoning-intensive tasks. The interaction between precedence constraints, task constraints, and weighted preferences induces a combinatorial search space that requires exact reasoning. While \LLM-based reasoning without external solvers (i.e., reasoning internally or via CoT chains) can produce explanations that appear convincing, such models provide no formal guarantees of feasibility or optimality, and optimisation errors are difficult to detect from the reasoning trace alone.

In contrast, when the same task is encoded as a Maximum Satisfiability (MaxSAT) problem by \GPT and solved using a symbolic solver, an optimal solution can be found reliably, and both feasibility and optimality can be easily verified. By prompting \LLMs to generate Python code that encodes the problem using a MaxSAT solver, the model produces the following optimal schedule:
\[
\{J_0:4, J_1:6, J_2:5, J_3:1, J_4:3, J_5:2\},
\]
which satisfies all hard constraints and minimises the total penalty induced by violated preferences.

This simple optimisation example illustrates a key insight motivating our work: \LLMs should not be trusted to perform optimisation reasoning, but they are highly effective at translating user intent into Python code. When reasoning is externalised from the \LLM and delegated to a sound and complete MaxSAT solver, the resulting system combines the strengths of both paradigms. The \LLM \emph{captures user constraints} and \emph{preferences} from natural language and generates an executable encoding, while the MaxSAT solver guarantees feasibility, optimality, and verifiability of the solution.
Rather than attempting to improve \LLMs’ internal reasoning capabilities for optimisation tasks, we treat them as \emph{interfaces between users and MaxSAT solvers}, enabling reliable, verifiable, and preference-aware decision-making grounded in formal guarantees.

% \section{Preference-Based Maximum Satisfiability}
\section{Preliminaries}
\label{sec:prelim}

The \emph{Boolean Satisfiability}~(SAT) problem is the canonical decision problem in propositional logic~\cite{biere2009handbook}. A \emph{literal} is defined as either a propositional variable $x_i$ or its negation $\neg x_i$. A formula is in \emph{Conjunctive Normal Form}~(CNF) if it is expressed as a conjunction of clauses, where each clause is a disjunction of literals.
Given a CNF formula $\phi$, the SAT problem consists of determining whether there exists a truth assignment to the variables that satisfies $\phi$, or whether no such assignment exists.
The \emph{Maximum Satisfiability}~(MaxSAT) problem extends SAT by introducing an optimisation objective~\cite{li2009maxsat,maxsat-handbook-sat}. Given a CNF formula $\phi$, the goal is to find an assignment that maximises the number of satisfied clauses.
In the \emph{partial MaxSAT} setting, the formula $\phi$ is partitioned into a set of hard clauses $\phi_h$ and a set of soft clauses $\phi_s$. The objective is to find an assignment that satisfies all hard clauses while minimising the number of unsatisfied soft clauses. In the \emph{weighted partial MaxSAT} variant, each soft clause is associated with a weight, and the objective becomes minimising the total weight of the unsatisfied soft clauses. In \emph{preference-based MaxSAT}~\cite{juma2011exploiting,DBLP:conf/ai/JumaHM12,DBLP:phd/ethos/Russell12}, user-defined preferences are modelled as soft constraints in a weighted partial MaxSAT formulation, where weights reflect the relative importance or ordering~of~user~preferences.
\begin{example}[Weighted Partial MaxSAT]
  Consider the partial MaxSAT formula $\phi = (\phi_h, \phi_s)$, where
  $\phi_h = \{ (x_1 \vee x_2), (\neg x_2 \vee x_3) \}$ and
  $\phi_s = \{ (\neg x_1), (\neg x_3) \}$, with weights
  $w = \{ w_{\neg x_1} = 1, \; w_{\neg x_3} = 3 \}$.
  The assignment $\{ (x_1, 1), (x_2, 0), (x_3, 0) \}$ is an optimal solution for $\phi$, as it satisfies all hard clauses in $\phi_h$ and violates only one soft clause in $\phi_s$, namely $(\neg x_1)$, incurring a total weight of~1.
  In contrast, the assignment $\{ (x_1, 0), (x_2, 1), (x_3, 1) \}$, while satisfying all hard clauses, violates the soft clause $(\neg x_3)$ and incurs a total weight of~3. Since this weight is higher than the optimal value of~1, the assignment is not optimal.
\end{example}
Moreover, given an assignment (possible solution) to the variables of a MaxSAT formula, its \emph{satisfiability} and \emph{optimality} can be \emph{verified} by adding the assignment as hard constraints to the solver and checking both the satisfiability of the resulting instance and the~associated~optimisation~cost.

\section{MaxSAT-Based Reasoning for \LLMs}
\label{sec:maxsat-based-reasoning-4-llms}

We propose a hybrid neuro-symbolic reasoning approach that combines the natural language understanding and code generation capabilities of Large Language Models~(\LLMs) with the formal and reliable optimisation guarantees of preference-based Maximum Satisfiability~(MaxSAT). The key idea is to \emph{externalise reasoning} from \LLMs into Python code and delegate automated reasoning to a MaxSAT solver, while independently verifying the correctness of the resulting solutions. All prompts used to interact with the models are provided in Appendix~\ref{sec:prompts}.

\subsection{Problem Setting}

We consider preference-based reasoning problems defined by:
(i) a set of \emph{hard constraints} that must be satisfied, and
(ii) a set of \emph{soft constraints} and/or preferences that should be optimised.
Such problems arise naturally in domains such as scheduling and robotics, where solutions must be feasible while optimising user-defined objectives. Each problem instance is provided to the model as a natural language description specifying constraints and preferences. The goal is to compute a solution that satisfies all hard constraints and is optimal with respect to the stated preferences.

%%%%%%%%%%%%%%%%%%%%%%%%%%%%%%%%%%%%%%%%%%%%%%%%%%%%%%%

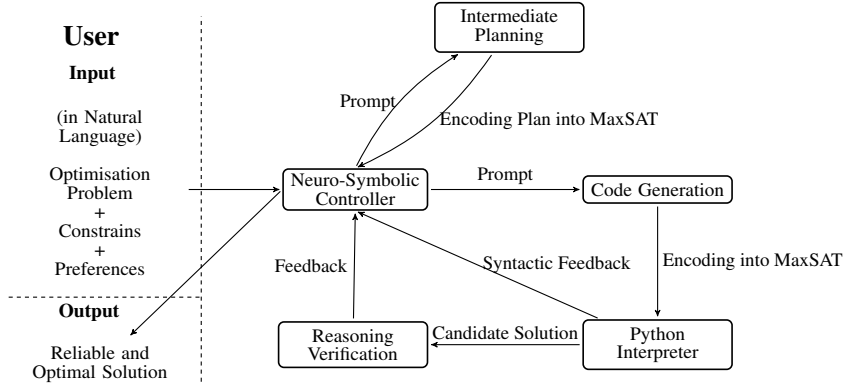
\begin{figure*}[t!]
\centering
% \vspace{2cm}
\resizebox{0.8\columnwidth}{!}{
\begin{tikzpicture}[node distance=2cm, auto,every node/.style={font=\fontsize{14}{14}\selectfont}]
 \node[punkt] (enum) {Neuro-Symbolic Controller};
 \node[punkt, inner sep=5pt, above=3cm of enum, xshift=4cm]
 (llm) {Intermediate Planning}
 edge[pil, <-, bend left=-15] node[left=-0.05cm]  (l) {Prompt} (enum.north)
 edge[pil, ->, bend left=15] node[right=0.05cm]  (l) {Encoding Plan into MaxSAT} (enum.north);
 \node[punkt, inner sep=5pt,right=4cm of enum]
 (inter) {Code Generation}
 edge[pil, <-] node[above]  (l) {Prompt} (enum.east);
 % edge[pil, ->, bend left=15] node[below]  (l) {Encoding using \texttt{PySAT}} (enum.east);
 \node[punkt, inner sep=5pt,below=3cm of inter]
 (veri) {Python\\Interpreter}
 edge[pil, <-] node[right]  (l) {Encoding into MaxSAT} (inter.south)
 edge[pil, ->] node[right] {Syntactic Feedback} (enum.south);
 \node[punkt, inner sep=5pt,left=4cm of veri]
 (fl) {Reasoning Verification}
 edge[pil, <-] node[above] {Candidate Solution} (veri.west)
 edge[pil] node[left=0.05cm] {Feedback} (enum.south);
\node[ionode,draw=none,left=2.5cm of enum, yshift=0cm] (specs_node) {\\(in Natural Language)\\\ \\Optimisation Problem\\+\\Constrains\\+\\Preferences}
  edge[pil] (enum.west);
\node[ionode,draw=none,below=1.5cm of specs_node] (output) {Reliable and Optimal Solution}
  edge[pil, <-] (enum.west);

% Vertical dashed division line
\draw[dashed, thick] (-4,-5) -- (-4,4.5);

\draw[dashed, thick] (-9,-2.8) -- (-4,-2.8);

% Labels
\node[anchor=east] at (0,4) {\textbf{\huge{}}};
\node[anchor=east] at (-6,4) {\textbf{\huge{User}}};
\node[anchor=east] at (-6.1,3) {\Large{\textbf{Input}}};
\node[anchor=east] at (-6,-3.2) {\textbf{\Large{Output}}};
% \node[anchor=east] at (4,3) {\textbf{\huge{Programmer}}};
\end{tikzpicture}
}
\caption{Our pipeline for enabling \LLMs to externalise reasoning through vibe coding, delegating formal reasoning and optimisation to a MaxSAT solver via its Python API. \LLMs' solutions are verified against a canonical MaxSAT encoding.}
\label{fig:approach}
\end{figure*}

\subsection{LLM-to-MaxSAT Pipeline}

Figure~\ref{fig:approach} illustrates our reasoning pipeline. Given a natural language problem description, we apply the following steps.

\paragraph{Intermediate Planning.}
We introduce an explicit intermediate \emph{planning} step. Before generating code, the \LLM is asked to produce a structured, symbolic description of the variables, constraints, and preferences. This step is inspired by prior work~\cite{DBLP:conf/naacl/HaoCZF25}, which shows that asking the model to first outline how to encode the problem can lead to improved results. The resulting plan is then provided back to the model as guidance for code generation. While not required for correctness, this step can improve modelling accuracy by encouraging more systematic encodings.

\paragraph{Code Generation.}
Rather than asking the \LLM to directly produce a solution, we prompt it to generate executable Python code that encodes the problem as a MaxSAT instance.
If the intermediate planning step is not performed, the \LLM begins by interpreting the user-provided natural language description, identifying the decision variables, hard constraints, and preferences. 
We explicitly instruct the model to use \texttt{PySAT}~\cite{imms18-PySAT}, a Python API for interacting with SAT and MaxSAT solvers.
The generated Python code constructs a weighted CNF formula in which: (i) hard constraints are encoded as hard clauses, and (ii) preferences are encoded as weighted soft clauses. Moreover, at this stage we provide a brief explanation and an example illustrating how to use the \texttt{PySAT} API, in order to familiarise the model with the interface and reduce the likelihood of errors in the generated code.

\paragraph{MaxSAT Solving.}
The generated code, which encodes the problem as a MaxSAT instance, invokes a MaxSAT solver. This guarantees that, whenever a solution exists, the returned assignment is feasible and optimal with respect to the encoded preferences, or at least with respect to the way the model has encoded the given problem. At this stage, the correctness and optimality of the solution no longer depend on the \LLM, but on the MaxSAT solver.

\paragraph{Verification Against Canonical Encodings.}
A central challenge in evaluating \LLM-generated encodings is that different encodings may yield different models or objective values, even when they represent the same underlying problem. To address this, we decouple \emph{solution validity} from \emph{encoding correctness} during \emph{evaluation}.
For each problem instance, we construct a \emph{canonical MaxSAT encoding} that serves as a reference specification only for evaluation purposes. This canonical encoding is not required by the proposed reasoning pipeline and is not assumed to be available in real-world deployments. Its sole purpose is to provide a model-independent ground truth that enables fair comparison across different \LLMs, prompting strategies, and encodings.
During the code generation step, we explain to the \LLM the \emph{exact format} expected for the solution. After executing the \LLM-generated Python code, we parse the solution returned by the solver and verify it against the canonical encoding by checking: (i) \emph{feasibility}, i.e., whether the solution satisfies all hard constraints; and (ii) \emph{optimality}, i.e., whether the solution achieves the same objective value as the canonical optimal solution. \textbf{This verification procedure allows multiple optimal solutions and alternative encodings to be accepted}, while ensuring that all accepted solutions are correct with respect to the reference semantics.
Importantly, this evaluation protocol does not rely on trusting the \LLM’s internal reasoning or reported objective values. 
In practical settings, correctness and optimality are instead guaranteed by the MaxSAT solver with respect to the constraints and preferences encoded by the \LLM, potentially complemented by interactive user refinement rather than by a hidden canonical~specification.

\paragraph{Feedback.}
If the generated code fails to execute, violates the required output format, or triggers a runtime/API error, we return the corresponding error message to the model and request a refined implementation. In our experiments, canonical feasibility and optimality checks are used only for final evaluation and are not returned to the model as feedback. Thus, refinement feedback does not expose the reference encoding or canonical objective value.
This feedback loop continues until the model produces a feasible and optimal solution, a predefined time limit is reached, or the number of iterations exceeds five, which typically indicates that the model is stuck and repeatedly generating the same incorrect solution.

Thus, by combining \LLM-driven semantic interpretation of user constraints and preferences with \LLM-generated code, exact MaxSAT solving, and independent verification, we transform unreliable natural language reasoning into a robust and verifiable optimisation process. This separation of concerns allows \LLMs to focus on understanding user intent, while formal reasoning and correctness are ensured by the MaxSAT solver.

\section{Experiments}
\label{sec:results}

The goal of our experiments was to evaluate Large Language Models~(\LLMs) on a range of reasoning tasks that require optimising an objective, by asking the models to encode these optimisation problems as MaxSAT instances via the generation of Python code. Accordingly, our experiments were designed to address the following research questions~(RQs):
\\
\textbf{RQ1.} Are \LLMs effective at encoding optimisation problems into MaxSAT formulations through \emph{vibe coding}?
\\
\textbf{RQ2.} Do \LLMs exhibit different performance across different families of optimisation problems?
\\
\textbf{RQ3.} To what extent does an intermediate planning step affect the quality of \LLM-generated Python code for MaxSAT-based optimisation?
\\
\textbf{RQ4.} How do different user preferences influence \LLMs’ reasoning behaviour and solution~quality?
\\
\textbf{RQ5.}~Do~\LLMs~perform better when leveraging plans generated~by~other~models?

\subsection{Experimental Setup}
All experiments were run using NVIDIA A100-SXM4 graphics card with 80GB of memory and 256GB of RAM, using a time limit of 5 minutes. For the MaxSAT oracle, RC2~\cite{imms19-RC2} from the \texttt{PySAT} toolkit~\cite{imms18-PySAT}~was~used, since \texttt{PySAT} is a Python API for several SAT/MaxSAT algorithms.

\subsection{Evaluation Dataset}
\label{sec:dataset}

To evaluate the proposed approach, we constructed a dataset of preference-based reasoning problems with natural language descriptions and reference encodings. The dataset is designed to assess a model’s ability to interpret user constraints and preferences, externalise reasoning through code generation, and produce solutions that are feasible and optimal with respect to a canonical specification.
\textbf{Problem Families:}
The dataset comprises three families of constraint optimisation problems typically used in MaxSAT competitions:
(1) \emph{Maximum Independent Set~(MIS)}, where the objective is to select the largest subset of mutually non-adjacent vertices;
(2) \emph{Scheduling}, where tasks must be assigned start times subject to precedence and resource constraints while optimising preference-based objectives; and
(3) \emph{Set Cover}, where a collection of sets must be selected to cover all required elements while minimising preference-based costs.
These problem families were chosen because they naturally involve a combination of hard constraints and soft preferences, and thus serve as a suitable proof of concept for this study.
The dataset contains 25 instances of varying size and difficulty for each problem family, enabling evaluation across different reasoning and modelling challenges. In total, the dataset includes four preference variants for each problem instance (namely, \texttt{none}, \texttt{p1}, \texttt{p2}, and \texttt{p3}), each accompanied by a natural language description, a canonical MaxSAT encoding, and a verified optimal solution. As a result, the dataset comprises \textbf{300 reasoning instances} in total.
\textbf{Natural Language Descriptions:}
Each instance is associated with a natural language description specifying: (i) the decision variables, (ii) the hard constraints that must be satisfied, and (iii) the preferences defining the optimisation objective. The descriptions are written to resemble user-provided specifications rather than formal problem definitions, requiring models to infer intent from natural~language.
\textbf{Canonical MaxSAT Encodings:}
For each instance, we provide a \emph{canonical preference-based MaxSAT formula} that serves as the reference specification for our evaluation. This encoding captures the intended semantics of the problem, with hard clauses representing mandatory constraints and weighted soft clauses representing preferences. The canonical encoding is solved using an exact MaxSAT solver to compute an optimal objective value. Importantly, the canonical encoding is used \emph{only for verification and evaluation}; models are not given access to this encoding or its solution during inference. During evaluation, canonical verification is used only to score the final output and is not used to provide feedback to the model.
Note that this evaluation protocol allows different encodings and multiple optimal solutions to be accepted, while ensuring that all reported successes correspond to correct solutions under the reference semantics.

\subsection{Large Language Models (\LLMs)}
In our evaluation, we consider four \LLMs: two open-source models and two closed-access models.
For the open-source models, we select \LLMs available on Hugging Face~\cite{huggingface}: Alibaba’s \Qwen~\cite{yang2025qwen3} (30B) and Meta’s \Llama~\cite{llama3.3} (70B). These specific model sizes were chosen as they correspond to the most recent and largest available versions of each model that can be run on the GPU infrastructure used in our experiments.
For the closed-access models, we evaluate OpenAI’s \GPT~\cite{chatgpt5.5} and Google’s \Gemini~\cite{gemini3.1}, as these are among the most widely adopted state-of-the-art proprietary models.
Moreover, to ensure consistency across experiments, the temperature of all models was set to zero.

\begin{table*}[t!]
\centering
\caption{Portion of feasible and optimal solution achieved by each \LLM, across all preferences (acceptance rate \%).}
\resizebox{0.85\columnwidth}{!}{
\begin{tabular}{llccccc}
\toprule
Family & Model & direct-answer & cot-answer & pot-answer & maxsat-no-plan & maxsat-with-plan \\
\midrule
\multirow{3}{*}{mis} & \Gemini & 0.0 & 4.0 & 11.0 & 36.0 & \textbf{56.0} \\
 & \GPT & 0.0 & 2.0 & 13.0 & 32.0 & \textbf{51.0} \\
 & \Llama & 0.0 & 1.0 & 5.0 & 18.0 & \textbf{24.0} \\
 & \Qwen & 0.0 & 0.0 & 6.0 & 15.0 & \textbf{20.0} \\
\midrule
\multirow{3}{*}{scheduling} & \Gemini & 0.0 & 1.0 & 14.0 & 44.0 & \textbf{59.0} \\
 & \GPT & 0.0 & 0.0 & 9.0 & 41.0 & \textbf{56.0} \\
 & \Llama & 0.0 & 0.0 & 3.0 & \textbf{9.0} & 8.0 \\
 & \Qwen & 0.0 & 0.0 & 0.0 & \textbf{22.0} & 4.0 \\
\midrule
\multirow{3}{*}{setcover} & \Gemini & 34.0 & 36.0 & 46.0 & 79.0 & \textbf{87.0} \\
 & \GPT & 29.0 & 33.0 & 45.0 & 76.0 & \textbf{82.0} \\
 & \Llama & 14.0 & 37.0 & 41.0 & \textbf{62.0} & 45.0 \\
 & \Qwen & 17.0 & 27.0 & 25.0 & \textbf{49.0} & 29.0 \\
\midrule
\bottomrule
\end{tabular}
}
\label{tab:aggregated-results}
\end{table*}

\begin{table*}[t]
\centering
\setlength{\tabcolsep}{0.5mm}
\caption{Acceptance rate (\%) by preference (none / p1 / p2 / p3).}
\resizebox{1\columnwidth}{!}{
\begin{tabular}{llccccc}
\toprule
Family & Model & direct-answer & cot-answer & pot-answer & maxsat-no-plan & maxsat-with-plan \\
\midrule
\multirow{3}{*}{mis} 
 & \Gemini & 0.0 / 0.0 / 0.0 / 0.0 & 16.0 / 0.0 / 0.0 / 0.0 & 40.0 / 4.0 / 0.0 / 0.0 & 100.0 / 4.0 / 20.0 / 20.0 & \textbf{100.0 / 44.0 / 40.0 / 40.0} \\
 & \GPT & 0.0 / 0.0 / 0.0 / 0.0 & 8.0 / 0.0 / 0.0 / 0.0 & 52.0 / 0.0 / 0.0 / 0.0 & 100.0 / 4.0 / 12.0 / 12.0  & \textbf{96.0 / 72.0 / 28.0 / 8.0} \\
 & \Llama & 0.0 / 0.0 / 0.0 / 0.0 & 4.0 / 0.0 / 0.0 / 0.0  & 20.0 / 0.0 / 0.0 / 0.0 & 64.0 / 8.0 / 8.0 / 0.0 & \textbf{60.0 / 12.0 / 12.0 / 12.0} \\
 & \Qwen & 0.0 / 0.0 / 0.0 / 0.0 & 0.0 / 0.0 / 0.0 / 0.0 & 12.0 / 4.0 / 4.0 / 4.0  & 20.0 / 16.0 / 12.0 / 12.0 & \textbf{12.0 / 68.0 / 0.0 / 0.0} \\
\midrule
\multirow{3}{*}{scheduling} 
 & \Gemini & 0.0 / 0.0 / 0.0 / 0.0 & 4.0 / 0.0 / 0.0 / 0.0 & 56.0 / 0.0 / 0.0 / 0.0  & 60.0 / 40.0 / 40.0 / 36.0 & \textbf{96.0 / 56.0 / 48.0 / 36.0} \\
 & \GPT & 0.0 / 0.0 / 0.0 / 0.0 & 0.0 / 0.0 / 0.0 / 0.0 & 36.0 / 0.0 / 0.0 / 0.0 & 56.0 / 80.0 / 16.0 / 12.0 & \textbf{92.0 / 92.0 / 20.0 / 20.0} \\
 & \Llama & 0.0 / 0.0 / 0.0 / 0.0 & 0.0 / 0.0 / 0.0 / 0.0 & 12.0 / 0.0 / 0.0 / 0.0 & \textbf{12.0 / 12.0 / 12.0 / 0.0} & 12.0 / 12.0 / 8.0 / 0.0 \\
 & \Qwen & 0.0 / 0.0 / 0.0 / 0.0 & 0.0 / 0.0 / 0.0 / 0.0 & 0.0 / 0.0 / 0.0 / 0.0 & \textbf{76.0 / 8.0 / 4.0 / 0.0} & 16.0 / 0.0 / 0.0 / 0.0 \\
\midrule
 \multirow{3}{*}{setcover} 
 & \Gemini & 88.0 / 12.0 / 16.0 / 20.0 & 92.0 / 16.0 / 16.0 / 20.0 & 100.0 / 28.0 / 28.0 / 28.0 & 100.0 / 84.0 / 60.0 / 72.0 & \textbf{100.0 / 88.0 / 80.0 / 80.0} \\
 & \GPT & 92.0 / 12.0 / 12.0 / 0.0 & 96.0 / 12.0 / 12.0 / 12.0 & 100.0 / 20.0 / 32.0 / 28.0  & 96.0 / 84.0 / 60.0 / 64.0 & \textbf{100.0 / 88.0 / 72.0 / 68.0} \\
 & \Llama & 44.0 / 4.0 / 4.0 / 4.0 & 100.0 / 16.0 / 16.0 / 16.0 & 100.0 / 16.0 / 24.0 / 24.0 & \textbf{92.0 / 72.0 / 44.0 / 40.0} & 68.0 / 60.0 / 32.0 / 20.0 \\
  
 & \Qwen & 68.0 / 0.0 / 0.0 / 0.0 & 96.0 / 4.0 / 4.0 / 4.0 & 96.0 / 4.0 / 0.0 / 0.0 & \textbf{80.0 / 56.0 / 40.0 / 20.0} & 64.0 / 24.0 / 20.0 / 8.0 \\
\midrule
\bottomrule
\end{tabular}
}
\label{tab:pref-results}
\end{table*}

\subsection{Evaluation}

\textbf{Baselines.}
We compare our approach against three commonly used prompting strategies: (1)~\textbf{Direct Answer}, where the \LLMs generate a solution directly from a natural language description; (2)~\textbf{Chain-of-Thought}~(CoT), where the model is prompted to reason step by step in natural language before producing a solution; and (3)~\textbf{Program-of-Thought}~(PoT), where the model generates Python code to derive a solution. 
Across these baselines, the model is responsible either for performing reasoning and optimization internally (Direct Answer and CoT) or for delegating them to generated Python code~(PoT), without access to formal solvers or external verification. Appendix~\ref{sec:prompts} provides the prompt templates used for all baselines.

\textbf{Configurations.}
We consider two configurations of our neuro-symbolic approach:
(1)~\textbf{MaxSAT~(no plan)}, in which the \LLM directly generates Python code encoding the optimisation problem as a MaxSAT problem; and
(2) \textbf{MaxSAT~(with plan)}, in which the \LLM first produces an intermediate plan of variables, constraints, and preferences before generating the corresponding~MaxSAT~encoding.

\textbf{Evaluation Metric.}
All \LLM-generated solutions are evaluated using feasibility and optimality verification against a canonical MaxSAT encoding, independent of the model’s internal reasoning or reported objective values.
A solution is counted as \emph{accepted} if and only if it satisfies all hard constraints and achieves an optimal objective value under the canonical encoding. We report \emph{acceptance rate}~(\%) as the proportion of instances for which a feasible and optimal solution is achieved.

\textbf{RQ1.}
Table~\ref{tab:aggregated-results} reports global acceptance rates across all preference variants. Direct-answer and CoT baselines perform poorly, especially on \emph{MIS} and \emph{scheduling}: direct-answer achieves \(0\%\) for all models, while CoT reaches at most \(4\%\) on \emph{MIS} and \(1\%\) on \emph{scheduling}. PoT improves slightly but remains far below MaxSAT-based methods. In contrast, MaxSAT with planning achieves \(56\%\) and \(51\%\) on \emph{MIS} for \Gemini and \GPT, and \(59\%\) and \(56\%\) on \emph{scheduling}, respectively. On \emph{set cover}, MaxSAT-based methods also obtain the best results, reaching \(87\%\) for \Gemini and \(82\%\) for \GPT. These results show that \LLMs can effectively encode optimisation problems into MaxSAT through \emph{vibe coding}, provided that optimisation is externalised to a symbolic solver.

\textbf{RQ2.}
Performance differs substantially across problem families. \emph{Set cover} is consistently the easiest: even prompting baselines obtain non-zero acceptance rates, and MaxSAT-based approaches achieve the highest overall results. By contrast, \emph{MIS} and \emph{scheduling} are more challenging, with prompting-only methods almost entirely failing and MaxSAT performance becoming more model-dependent. This suggests that \LLM-generated MaxSAT encodings are sensitive to problem structure: coverage-style constraints are easier to encode than problems requiring precise modelling of exclusivity, precedence, feasibility, and combinatorial interactions.

\textbf{RQ3.}
Table~\ref{tab:aggregated-results} also compares MaxSAT encodings with and without intermediate planning. Planning consistently improves performance for the stronger models. For \Gemini, acceptance increases from \(36\%\) to \(56\%\) on \emph{MIS}, \(44\%\) to \(59\%\) on \emph{scheduling}, and \(79\%\) to \(87\%\) on \emph{set cover}; \GPT shows a similar trend, improving from \(32\%\) to \(51\%\), \(41\%\) to \(56\%\), and \(76\%\) to \(82\%\), respectively. However, planning is not uniformly beneficial: for \Llama and \Qwen, it sometimes degrades performance, particularly on \emph{scheduling} and \emph{set cover}. Thus, planning helps when the model can translate the plan into correct executable MaxSAT code, but may otherwise introduce~error~propagation.

\textbf{RQ4.}
Table~\ref{tab:pref-results} shows that preferences substantially affect solution quality. The \texttt{none} setting is generally easiest, while preference variants often reduce acceptance rates. However, the degradation is not strictly monotonic: some preferences appear qualitatively harder than others. For example, \Qwen with MaxSAT and planning reaches \(68\%\) on \emph{MIS} under \texttt{p1}, but \(0\%\) under both \texttt{p2} and \texttt{p3}. Despite these drops, MaxSAT-based methods remain more robust than direct-answer, CoT, and PoT baselines, especially for \emph{set cover}, where \Gemini and \GPT maintain strong performance even under \texttt{p2} and \texttt{p3}. Overall, user preferences significantly influence \LLM reasoning and encoding quality, but MaxSAT mitigates much of this difficulty.

\textbf{RQ5.}
Table~\ref{tab:maxsat-with-prev-plan-from} evaluates whether models benefit from plans generated by other models, focusing on \emph{MIS}; extended results for all families are reported in Table~\ref{tab:maxsat-with-prev-plan-from-2} in Appendix~\ref{sec:maxsat-with-other-plans}. Cross-model plans can improve individual cases: for example, \Gemini improves on \texttt{p1} and \texttt{p2} using \Llama- and \GPT-generated plans, while \GPT achieves its best \texttt{p1} and \texttt{p2} results with \Llama- and \Gemini-generated plans. However, these gains are inconsistent, especially for weaker models, where transferred plans often lead to unstable or near-zero performance. The appendix results show the same pattern across \emph{scheduling} and \emph{set cover}: transfer is most useful for simpler, more modular encodings such as \emph{set cover}, but remains brittle for more constrained families. Overall, cross-model CoTs or plans are useful only when the transferred plan is clear and the target model can faithfully implement it as executable MaxSAT code.

% ===== maxsat-with-prev-plan-from =====
\begin{table}[t]
\centering
\setlength{\tabcolsep}{1.5mm}
\caption{Acceptance rates when \LLMs generate MaxSAT encodings using an intermediate planning step produced either by the same model or by a different model.}
\resizebox{0.65\columnwidth}{!}{
\begin{tabular}{lllcccc}
\toprule
Family & Model & Plan-from & none & p1 & p2 & p3 \\
\midrule
  \multirow{9}{*}{mis}
 & \multirow{3}{*}{\Gemini}  & \Gemini &  \textbf{100.0} & 44.0 &  40.0 &  \textbf{40.0} \\
 &  & \GPT & \textbf{100.0} & 72.0 &  \textbf{48.0} & 0.0 \\
 &  & \Llama & 88.0 & \textbf{84.0} & 8.0 & 4.0 \\
 &  & \Qwen & 96.0 & 72.0 & 0.0 & 8.0 \\
\cmidrule(lr){2-7}  
 & \multirow{3}{*}{\GPT} & \Gemini & 96.0 & 72.0 & \textbf{32.0} & \textbf{8.0} \\
 &  & \GPT & 96.0 & 72.0 & 28.0 & \textbf{8.0} \\
 &  & \Llama & 88.0 & \textbf{84.0} & 8.0 & 4.0 \\
 & & \Qwen & \textbf{100.0} & 72.0 & 4.0 & 0.0 \\
\cmidrule(lr){2-7}
 & \multirow{3}{*}{\Llama} & \Gemini & \textbf{68.0} & 12.0 & \textbf{12.0} & \textbf{12.0} \\ 
 &  & \GPT & 0.0 & 24.0 & 4.0 & 0.0 \\
 &  & \Llama & 60.0 & 12.0 & \textbf{12.0} & \textbf{12.0} \\ 
 &  & \Qwen & 0.0 & \textbf{68.0} & 0.0 & 0.0 \\
 \cmidrule(lr){2-7}
 & \multirow{3}{*}{\Qwen} & \Gemini & \textbf{12.0} & \textbf{68.0} & 0.0 & \textbf{8.0} \\
 &  & \GPT & 0.0 & 16.0 & 0.0 & 0.0 \\
 &  & \Llama & 4.0 & 52.0 & 0.0 & 0.0 \\
 & & \Qwen & \textbf{12.0} & \textbf{68.0} & 0.0 & 0.0 \\
\bottomrule
\end{tabular}
}
\label{tab:maxsat-with-prev-plan-from}
\end{table}

% \subsection{Discussion}
% \label{sec:discussion}

\paragraph{Discussion.} Our evaluation yields four main findings. First, consistent with prior work~\cite{pan2023logic,ye2023satlm,haoplanning}, \LLMs are largely ineffective at solving constrained optimisation problems through internal reasoning alone: direct-answer and CoT prompting almost entirely fail on \emph{MIS} and \emph{scheduling}, while PoT provides only limited gains.
Second, encoding optimisation problems into MaxSAT via Python code generation substantially improves acceptance rates, especially for stronger models such as \Gemini and \GPT. This supports our central claim that \emph{externalising optimisation reasoning to a symbolic MaxSAT solver} enables more reliable and verifiable preference-aware reasoning than relying on the \LLM alone.
Third, intermediate planning helps only when the model can translate the plan into correct executable code. While planning consistently improves MaxSAT performance for \Gemini and \GPT, it can degrade results for \Llama and \Qwen, indicating that the planning step may also introduce error propagation.
Finally, cross-model plans show limited and task-dependent transferability. They can improve selected cases, particularly for simpler problems such as \emph{set cover}, but may hurt performance when the transferred plan is misaligned with the target model's encoding strategy. Overall, the most reliable configuration is not the one that produces more reasoning text, but the one that produces executable, solver-verifiable MaxSAT~encodings.

\section{Related Work}
\label{sec:related}

\emph{Neuro-symbolic reasoning methods.}
Recent work has explored neuro-symbolic approaches for improving \LLM reasoning across planning~\cite{guan2023leveraging,valmeekam2023planning,kambhampati2024position,DBLP:conf/naacl/HaoCZF25,haoplanning}, arithmetic~\cite{gao2023pal}, and logical reasoning~\cite{xu2025aristotle,calanzone2024logically,linzebralogic,ye2023satlm,pan2023logic,shi2025constraintllm,szeider2024mcp}. Representative systems such as \texttt{SATLM}~\cite{ye2023satlm} and \texttt{LOGIC-LM}~\cite{pan2023logic} combine \LLMs with symbolic solvers by translating natural language problems into formal logical specifications, which are then solved by SMT solvers or other deterministic inference engines. By separating parsing from reasoning and execution, these methods reduce errors commonly observed in chain-of-thought~(CoT) prompting; however, directly generating logical formulae can still produce syntactically invalid or ill-formed encodings. 
More recently, \texttt{ConstraintLLM}~\cite{shi2025constraintllm} introduced a neuro-symbolic framework for constraint programming that combines a fine-tuned \LLM with retrieval, tree-of-thought exploration, and solver-guided self-correction to generate and validate constraint models at industrial scale. Similarly, work on realistic travel-planning benchmarks shows that \LLMs struggle to generate valid plans directly for complex multi-constraint problems, but become more reliable when guided to formalise queries and invoke SMT-based verification tools~\cite{DBLP:conf/naacl/HaoCZF25}. \texttt{LLMFP}~\cite{haoplanning} generalises this paradigm to diverse planning domains by framing problems as constrained optimisation tasks solved with SMT solvers. Like these approaches, our work externalises reasoning by using \LLMs to \emph{vibe code} constraints into executable Python. However, prior work primarily focuses on feasibility or planning with SMT-based APIs, whereas we target preference-aware optimisation using \texttt{PySAT}~\cite{imms18-PySAT} and SAT/MaxSAT solvers.

\emph{Preference-based MaxSAT.}
In preference-based MaxSAT~\cite{juma2011exploiting,DBLP:conf/ai/JumaHM12,DBLP:phd/ethos/Russell12}, user preferences are represented as soft constraints in a weighted partial MaxSAT formulation, where weights encode their relative importance or priority. Existing approaches typically require users to specify constraints and preferences in the Planning Domain Definition Language~(PDDL)~\cite{aeronautiques1998pddl}. While this provides an explicit and rigorous representation of user intent, it also requires familiarity with a specialised domain-specific language.

Our work differs from prior hybrid \LLM--solver approaches in three main ways. First, we address \emph{preference-based optimisation} rather than only feasibility checking or single-objective planning. Second, we use weighted soft constraints and independently verify optimality, rather than relying on the \LLM's internal reasoning or encoding. Third, the optimisation problems considered here are predominantly Boolean, making SAT/MaxSAT a natural formalism; SMT-based methods typically reduce such Boolean structure to SAT internally.
To the best of our knowledge, this is the first work to combine \LLMs for extracting user intent from natural language with MaxSAT solvers for rigorous preference-aware optimisation under user-defined constraints.

\section{Conclusion}
\label{sec:conclusion}

We studied the use of Large Language Models~(\LLMs) for preference-aware optimisation by externalising reasoning to Maximum Satisfiability~(MaxSAT) solvers. Our results show that, while \LLMs are unreliable at solving optimisation problems through internal reasoning alone, they can effectively translate natural language constraints and preferences into executable MaxSAT encodings via Python code generation. By delegating optimisation and verification to symbolic solvers, our approach enables more reliable and verifiable preference-aware reasoning.
Across multiple optimisation families, our evaluation shows that solution quality depends on problem structure, preference complexity, and intermediate planning. Overall, these findings highlight the value of separating semantic interpretation from formal optimisation, and suggest that combining \LLMs with MaxSAT solvers is a promising direction for robust user-facing decision-making systems.
Future work will explore interactive optimisation settings in which \LLMs and logical reasoners support user-in-the-loop decision making, enabling users to iteratively refine preferences and request revised solutions as requirements~evolve.

\section*{Acknowledgements}
This work was supported by grant PID2022-139835NB-C21 funded by MCIN/\-AEI/\-10.13039/\-501100011033. 
This project has received funding from the European Union’s HORIZON-MSCA-2025-PF research and innovation programme under the Marie Skłodowska-Curie  (Sherlock4Py,~GA~No~101269051).
This project was additionally supported by the ALLIES Cofund, and has received funding from the European Union’s Horizon-MSCA-2022-COFUND-01 research and innovation programme under the Marie Skłodowska-Curie~(GA~No~101126626). 
This project also received funding from ELSA: European Lighthouse on Secure and Safe AI project~(GA~No.~101070617 under~UK~guarantee). PO acknowledges travel support from ELSA Mobility~Program~(GA~No~101070617).

% \section*{Impact Statement}

% This paper explores a neuro-symbolic approach for preference-aware optimisation that prioritises formal verification and correctness. By externalising reasoning from Large Language Models to sound and complete MaxSAT solvers, the approach aims to reduce errors and unreliable reasoning in optimisation tasks. While the work is methodological and not application-specific, its emphasis on verifiable reasoning aligns with broader efforts toward responsible and trustworthy AI. No immediate negative societal impacts are anticipated.


\bibliographystyle{plainnat}

\bibliography{bibliography}

%%%%%%%%%%%%%%%%%%%%%%%%%%%%%%%%%%%%%%%%%%%%%%%%%%%%%%%%%%%%%%%%%%%%%%%%%%%%%%%
%%%%%%%%%%%%%%%%%%%%%%%%%%%%%%%%%%%%%%%%%%%%%%%%%%%%%%%%%%%%%%%%%%%%%%%%%%%%%%%
% APPENDIX
%%%%%%%%%%%%%%%%%%%%%%%%%%%%%%%%%%%%%%%%%%%%%%%%%%%%%%%%%%%%%%%%%%%%%%%%%%%%%%%
%%%%%%%%%%%%%%%%%%%%%%%%%%%%%%%%%%%%%%%%%%%%%%%%%%%%%%%%%%%%%%%%%%%%%%%%%%%%%%%
\newpage
\appendix
\onecolumn

\section{Prompt Templates}
\label{sec:prompts}

This appendix summarises the prompt templates used to interact with the \LLMs across all experimental configurations. All prompts are executed with a temperature of~0 and are designed to explicitly separate semantic interpretation, planning, and formal optimisation. The prompts enforce solver-backed reasoning and avoid reliance on the \LLM’s internal reasoning~traces.

\subsection{Intermediate Planning Prompt}

When the intermediate planning step is enabled, the \LLM is first asked to produce a symbolic encoding plan before generating any code:

\begin{verbatim}
You are an expert MaxSAT/SAT encoding designer.

Given ONLY a natural-language problem description, draft a precise encoding plan
with the following sections:

1) Variables
2) Hard constraints
3) Soft constraints / preferences
4) Objective (weighted MaxSAT)
5) Output schema (repeat the JSON schema to emit)

Do NOT write code.
Do NOT use external data.
\end{verbatim}

The generated plan is  provided verbatim to the subsequent code-generation step.

\subsection{RC2 Solver Usage Prompt}
\label{sec:rc2-help}

The following prompt is provided verbatim to the \LLMs to explain how to use the RC2 MaxSAT solver through the \texttt{PySAT} API:

\begin{verbatim}
# RC2 / WCNF usage
# IMPORTANT: The call rc2.compute() returns the model (a list of integer literals).
#            If rc2.compute() returns None, then the WCNF formula is UNSAT.
#            The objective (weighted cost) is available as rc2.cost after compute().
#
# Use this exact pattern in your solve() implementation:
from pysat.formula import WCNF
from pysat.examples.rc2 import RC2

def rc2_solve_example():
    w = WCNF()
    # Example hard clause: w.append([1, -2])
    # Example soft clause with weight 3: w.append([-3], weight=3)

    rc2 = RC2(w)
    # rc2.compute() returns the model (list of ints) or None if the MaxSAT 
    # formula is UNSAT
    model_lits = rc2.compute()   # <- model_lits is a list of signed ints
    # if model_lits is None then the formula is UNSAT
    # Read the cost (sum of falsified soft clause weights) after compute():
    cost = int(rc2.cost)
    assignment = {abs(l): (l > 0) for l in model_lits}
    return cost, assignment

# End of RC2_HELP
\end{verbatim}

This prompt is prepended to the code-generation instructions to ensure consistent and correct interaction with the MaxSAT solver across all models and experimental configurations, and to reduce API misuse and syntactic errors in the generated code.

\subsection{Code Generation Prompt (with Plan)}

If a plan is available, the following user prompt is used to generate the MaxSAT encoding:

\begin{verbatim}
Natural-language problem description:
---
<problem description>
---

Encoding PLAN:
---
<generated plan>
---

You are a code model that writes standalone Python using PySAT to solve a 
planning/optimization task from a natural-language description.
Your job: Write PySAT code that exactly models the hard rules and preferences 
and computes the optimal solution.
You MUST use python-sat (PySAT) to build and solve a SAT/MaxSAT model 
matching the problem.
Use the RC2 example provided in RC2_HELP exactly as a template if helpful.
Write PySAT code implementing the PLAN. Return JSON ONLY as specified.
\end{verbatim}

\subsection{Code Generation Prompt (without Plan)}

If the intermediate planning step is disabled, the \LLM is prompted directly from the natural-language description:

\begin{verbatim}
Problem description:
---
<problem description>
---

You are a code model that writes standalone Python using PySAT 
to solve a planning/optimization task from a natural-language description.
Your job: Write PySAT code that exactly models the hard rules 
and preferences and computes the optimal solution.
You MUST use python-sat (PySAT) to build and solve a 
SAT/MaxSAT model matching the problem.
Use the RC2 example provided in RC2_HELP exactly as a template if helpful.
Return JSON ONLY as specified.
\end{verbatim}

\subsection{Direct-Answer Baseline Prompt}

For the direct-answer baseline, the \LLM is asked to produce a solution without generating code or invoking solvers:

\begin{verbatim}
Problem description:
---
<problem description>
---

You are an expert solver.

Your Job: Provide the optimal solution to the optimization task 
from a natural-language description, in using the required JSON schema.

Return ONLY the JSON. No code. JSON only.
\end{verbatim}

\subsection{Chain-of-Thought Baseline Prompt}

For the chain-of-thought (CoT) baseline, the \LLM is encouraged to reason step by step before producing the final answer:

\begin{verbatim}
Problem description:
---
<problem description>
---

You are an expert solver.

Your Job: Provide the optimal solution to the optimization task 
from a natural-language description, using the required JSON schema.

Think step-by-step to derive the optimal solution.

Write numbered steps. Then output the final JSON on the last line.
\end{verbatim}

During evaluation, only the final JSON output is considered; the reasoning trace itself is ignored.

\subsection{Program-of-Thought Baseline Prompt}

For the Program-of-Thought~(PoT) baseline, the \LLM is prompted to generate standalone Python code that computes the optimal solution without using any symbolic solver:

\begin{verbatim}
Problem description:
---
<problem description>
---

You are an expert Python programmer.

Your task is to solve the optimisation problem described above and
return the optimal solution using the required JSON schema.

Write a standalone Python script that exactly models the hard rules
and preferences and computes the optimal solution for the planning or
optimisation task.

You CANNOT use any symbolic solver or constraint satisfaction method.

Return JSON ONLY as specified.
\end{verbatim}

\subsection{Feedback Prompt for Iterative Refinement}

When the generated code fails to execute or produces a non-optimal solution, the following feedback prompt is used to guide refinement:

\begin{verbatim}
You previously provided a response for the problem below, but it did not produce
a correct optimal solution or failed to run.

Problem description:
---
<problem description>
---

Previous output or code:
---
<previous model output>
---

Execution result and errors:
---
<stderr and solver feedback>
---

Please provide a corrected solution.
Follow the exact same output specification as before.
If producing code, return a single Python code block that prints:
{ "objective_cost": <int>, "solution_json": <JSON> }
and nothing else.
\end{verbatim}

This feedback loop is repeated for up to five attempts per instance.

All prompts are designed to: (i) externalise optimisation reasoning from the \LLM; (ii) enforce solver-backed reasoning via MaxSAT; (iii) avoid reliance on the model’s internal reasoning traces; and (iv) enable independent verification of feasibility and optimality during evaluation.

\section{Limitations}
\label{sec:limitations}

In this section we present the possible limitations of our work. 
First, our evaluation focuses on three families of Boolean optimisation problems: maximum independent set, scheduling, and set cover. Although these families cover different constraint structures and preference types, they do not capture the full range of optimisation problems that may arise in real-world decision-making systems. In particular, problems involving continuous variables, rich numeric constraints, uncertainty, or multi-agent interaction may require formalisms beyond SAT/MaxSAT.
Second, our evaluation relies on canonical MaxSAT encodings to verify feasibility and optimality. These encodings are used only for offline evaluation and are not assumed to be available in deployment. In practical settings, the MaxSAT solver guarantees optimality only with respect to the constraints and preferences encoded by the \LLM. Therefore, while our approach makes optimisation solver-verifiable, it does not by itself guarantee that the generated encoding fully captures the user's intended semantics.
Third, performance remains model-dependent. Stronger models such as \Gemini and \GPT benefit substantially from MaxSAT-based code generation, while weaker models are less reliable, especially on more constrained problem families and harder preference variants. This indicates that solver-backed reasoning reduces, but does not eliminate, the need for accurate semantic parsing and modelling by the \LLM.
Finally, our benchmark consists of controlled problem instances with structured natural language descriptions. Real user inputs may be more ambiguous, underspecified, or inconsistent. Handling such cases may require interactive clarification, richer validation mechanisms, or human oversight. We view these directions as important future work for deploying \LLM--MaxSAT systems in practical preference-aware decision-making settings.

\newpage

\section{Leveraging Chains of Thought~(CoTs) generated by other models}
\label{sec:maxsat-with-other-plans}

Table~\ref{tab:maxsat-with-prev-plan-from-2} presents the complete acceptance rates for MaxSAT encodings generated using an intermediate planning step produced either by the same model or by a different model.

% ===== maxsat-with-prev-plan-from =====
\begin{table}[t]
\centering
\setlength{\tabcolsep}{1.5mm}
\caption{Acceptance rates when \LLMs generate MaxSAT encodings using an intermediate planning step produced either by the same model or by a different model.}
\resizebox{0.7\columnwidth}{!}{
\begin{tabular}{lllcccc}
\toprule
Family & Model & Plan-from & none & p1 & p2 & p3 \\
\midrule
  \multirow{9}{*}{mis}
 & \multirow{3}{*}{\Gemini}  & \Gemini &  \textbf{100.0} & 44.0 &  40.0 &  \textbf{40.0} \\
 &  & \GPT & \textbf{100.0} & 72.0 &  \textbf{48.0} & 0.0 \\
 &  & \Llama & 88.0 & \textbf{84.0} & 8.0 & 4.0 \\
 &  & \Qwen & 96.0 & 72.0 & 0.0 & 8.0 \\
\cmidrule(lr){2-7}  
 & \multirow{3}{*}{\GPT} & \Gemini & 96.0 & 72.0 & \textbf{32.0} & \textbf{8.0} \\
 &  & \GPT & 96.0 & 72.0 & 28.0 & \textbf{8.0} \\
 &  & \Llama & 88.0 & \textbf{84.0} & 8.0 & 4.0 \\
 & & \Qwen & \textbf{100.0} & 72.0 & 4.0 & 0.0 \\
\cmidrule(lr){2-7}
 & \multirow{3}{*}{\Llama} & \Gemini & \textbf{68.0} & 12.0 & \textbf{12.0} & \textbf{12.0} \\ 
 &  & \GPT & 0.0 & 24.0 & 4.0 & 0.0 \\
 &  & \Llama & 60.0 & 12.0 & \textbf{12.0} & \textbf{12.0} \\ 
 &  & \Qwen & 0.0 & \textbf{68.0} & 0.0 & 0.0 \\
 \cmidrule(lr){2-7}
 & \multirow{3}{*}{\Qwen} & \Gemini & \textbf{12.0} & \textbf{68.0} & 0.0 & \textbf{8.0} \\
 &  & \GPT & 0.0 & 16.0 & 0.0 & 0.0 \\
 &  & \Llama & 4.0 & 52.0 & 0.0 & 0.0 \\
 & & \Qwen & \textbf{12.0} & \textbf{68.0} & 0.0 & 0.0 \\
\midrule
% \cmidrule(lr){2-7}
\midrule
  \multirow{9}{*}{scheduling}
 & \multirow{3}{*}{\Gemini} & \Gemini & \textbf{96.0} & 56.0 & 48.0 & \textbf{36.0} \\
 &  & \GPT & \textbf{96.0} & 80.0 & \textbf{52.0} & 12.0 \\
 &  & \Llama & 56.0 & 64.0 & 4.0 & 12.0 \\
 &  & \Qwen & 48.0 & \textbf{92.0} & 12.0 & 16.0\\
\cmidrule(lr){2-7}
 & \multirow{3}{*}{\GPT} & \Gemini & \textbf{96.0} & 80.0 & 8.0 & \textbf{20.0} \\
 &  & \GPT & 92.0 & \textbf{92.0} & 20.0 & \textbf{20.0} \\
 &  & \Llama & 56.0 & 64.0 & 4.0 & 12.0 \\
 &   & \Qwen & 48.0 & \textbf{92.0} & \textbf{24.0} & 16.0 \\
\cmidrule(lr){2-7}
 & \multirow{3}{*}{\Llama} & \Gemini & \textbf{20.0} & \textbf{12.0} & 4.0 & \textbf{4.0} \\       
 &  & \GPT & 8.0 & 8.0 & 4.0 & \textbf{4.0} \\
 &  & \Llama & 12.0 & \textbf{12.0} & \textbf{8.0} & 0.0 \\
 &  & \Qwen & \textbf{20.0} & 0.0 & 0.0 & 0.0 \\
\cmidrule(lr){2-7}
 & \multirow{3}{*}{\Qwen} & \Gemini & \textbf{60.0} & \textbf{20.0} & 0.0 & 0.0 \\               
 &  & \GPT & 16.0 & 8.0 & 0.0 & 0.0 \\
 &  & \Llama & \textbf{60.0} & \textbf{20.0} & 0.0 & 0.0 \\
 &  & \Qwen & 16.0 & 0.0 & 0.0 & 0.0 \\
  
\midrule
  \multirow{9}{*}{setcover}
 & \multirow{3}{*}{\Gemini} & \Gemini & \textbf{100.0} & 88.0 & \textbf{80.0} & \textbf{80.0}\\ 
 &  & \GPT & \textbf{100.0} & 88.0 & 68.0 & 64.0 \\
 &  & \Llama & \textbf{100.0} & \textbf{96.0} & 72.0 & 60.0 \\
 &  & \Qwen & \textbf{100.0} & 92.0 & 68.0 & 60.0 \\ 
\cmidrule(lr){2-7}
 & \multirow{3}{*}{\GPT} & \Gemini & \textbf{100.0} & 88.0 & \textbf{72.0} & 64.0 \\
 &  & \GPT & \textbf{100.0} & 88.0 & \textbf{72.0} & \textbf{68.0} \\
 &  & \Llama & \textbf{100.0} & \textbf{96.0} & \textbf{72.0} & 60.0 \\
 &  & \Qwen & \textbf{100.0} & 92.0 & \textbf{72.0} & 60.0 \\
\cmidrule(lr){2-7}
 & \multirow{3}{*}{\Llama} & \Gemini & \textbf{80.0} & \textbf{72.0} & \textbf{52.0} & \textbf{36.0} \\  
 &  & \GPT & 68.0 & 20.0 & 12.0 & 24.0 \\
 &  & \Llama & 68.0 & 60.0 & 32.0 & 20.0 \\
 &  & \Qwen & 76.0 & \textbf{72.0} & \textbf{52.0} & \textbf{36.0} \\  
 \cmidrule(lr){2-7}
 & \multirow{3}{*}{\Qwen} & \Gemini & 72.0 & \textbf{60.0} & \textbf{44.0} & \textbf{36.0} \\
 &  & \GPT & \textbf{76.0} & 16.0 & 32.0 & \textbf{36.0} \\
 &  & \Llama & 72.0 & \textbf{60.0} & \textbf{44.0} & \textbf{36.0} \\
 &  & \Qwen & 64.0 & 24.0 & 20.0 & 8.0 \\
\midrule
\bottomrule
\end{tabular}
}
\label{tab:maxsat-with-prev-plan-from-2}
\end{table}

%%%%%%%%%%%%%%%%%%%%%%%%%%%%%%%%%%%%%%%%%%%%%%%%%%%%%%%%%%%%

\end{document}